\documentclass[letterpaper, 10 pt, conference]{ieeeconf} 
\usepackage{amsmath,amsfonts,amssymb}
\usepackage{graphicx}
\usepackage{multirow}
\IEEEoverridecommandlockouts                              % This command is only needed if 
\title{\LARGE \bf
Robust Meta Learning for Image based tasks
}

\author{Penghao Jiang, Ke Xin, Zifeng Wang, Chunxi Li\thanks{Authors are with The Australian National University, Canberra, Australia.
		The first two authors contributed equally as joint first authorship. The last
		two authors contributed equally as joint second authorship.}}%}

\begin{document}

\maketitle
\thispagestyle{empty}
\pagestyle{empty}

%%%%%%%%%%%%%%%%%%%%%%%%%%%%%%%%%%%%%%%%%%%%%%%%%%%%%%%%%%%%%%%%%%%%%%%%%%%%%%%%
\begin{abstract}

A machine learning model that generalizes well
should obtain low errors on unseen test examples. Thus, if
we learn an optimal model in training data, it could have
better generalization performance in testing tasks. However,
learn such model is not possible in standard machine learning
frameworks as the distribution of the test data is unknown. To
tackle this challenge, we propose a novel robust meta learning
method, which is more robust to the image-based testing tasks
which is unknown and has distribution shifts with traning tasks.
Our robust meta learning method can provide robust optimal
models even when data from each distribution are scarce. In
experiments, we demonstrate that our algorithm not only has
better generalization performance, but also robust to different
unknown testing tasks.

\end{abstract}

%%%%%%%%%%%%%%%%%%%%%%%%%%%%%%%%%%%%%%%%%%%%%%%%%%%%%%%%%%%%%%%%%%%%%%%%%%%%%%%%
\section{Introduction}
Deep learning achieves good performance in many realworld
tasks such as visual recognition \cite{20,11}, but
relies on large training data, which indicates it is unable to
generalize to small data regimes. To overcome this limitation
of deep learning, recently, researchers have explored metalearning
\cite{15,29} approaches, whose goal is to learn a
good meta-initialization by extracting common (or meta)
knowledge over the distribution of tasks, rather than instances
from a single task. One of the most popular approaches is
the optimization-based meta-learning because it is modelagnostic
and can be applied to various downstream tasks
such as few-shot learning \cite{9,26}, reinforcement learning
\cite{39} and federated learning \cite{4}.

Unfortunately, most existing optimization-based metalearning
methods \cite{9,26} have a highly restrictive assumption
that tasks come from the same distribution (indistribution,
ID). However, distribution shifts between tasks are
usually inevitable due to data selection biases or various
confounders (e.g, background, color or texture) that widely
exist in the real-world \cite{32}. As shown in Figure 1, the
robot is testing in the distribution different from the training
distribution, which is a realistic scenario where tasks come
from different distributions (out-of-distribution, OOD) \cite{41}.

In this paper, we propose Robust Meta Learning (RML)
for out-of-distribution tasks to improve the performance of
meta-learning on image-based tasks, which learn invariant
representation for data from different distribution. To address
the distribution shift issue, we consider a regularization into
the update process of the meta-learner so as to improve the
generalization ability, where the invariant representation is
learned with such regularization. Our contribution can be
summarized as follows:

\begin{figure}[t]
\centering
\includegraphics[width=\columnwidth]{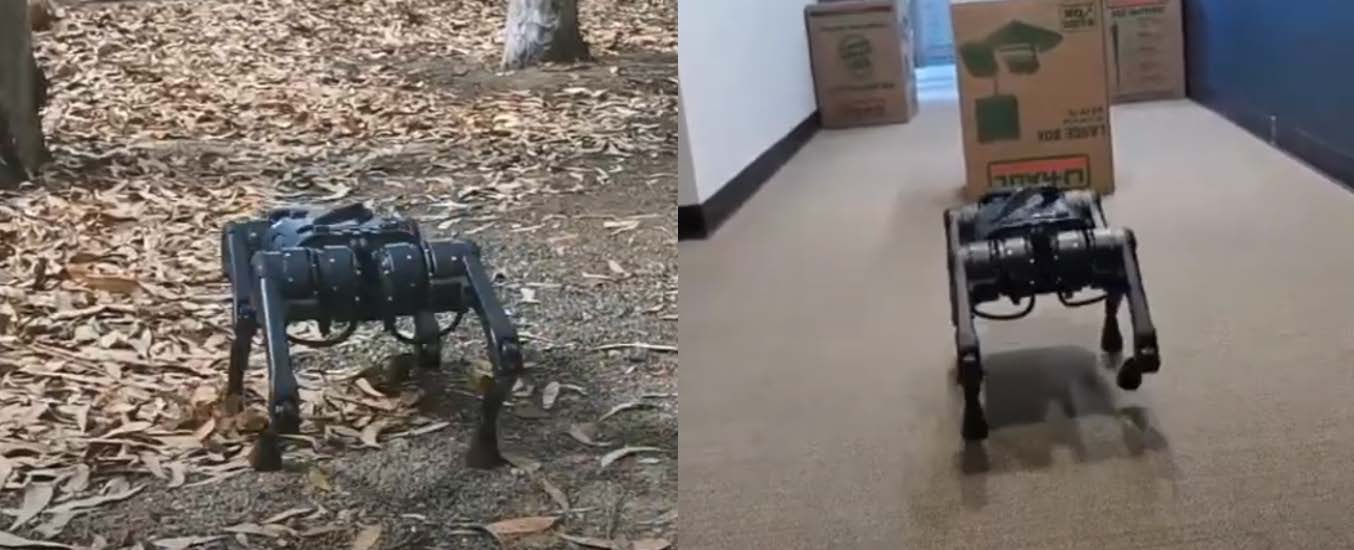}
\caption{Illustration examples of distribution shifts in imagebased tasks, where the robot is testing in the distribution
	different from the training distribution.}
\end{figure}

\begin{itemize}
\item First, we consider the problem caused by distribution
shift in image-based tasks, which has been less explored
in the literature;
\item Then, we propose Robust Meta Learning (RML) with a
novel regularization for out-of-distribution tasks, which
alleviates the domain shift between meta-training and
meta-test phase and thus improves the generalization;
\item Thirdly, we conduct extensive experiments to demonstrate
the effectiveness of our proposed method, in
comparison to state-of-the-art baselines.
\end{itemize}

\section{Related Work}
Meta-learning \cite{2,36}, also known as learning-to-learn,
is a learning framework devised for fast adaptation for
unseen tasks. Optimization-based meta-learning \cite{10} aims
to learn a general meta-learner, which can quickly adapt
to various task-specific base-learners for new tasks with a
few examples. LSTM meta-learner \cite{27} is proposed based
on the similarity between gradient-based update and cellstate
update in LSTM. The famous MAML \cite{10} learns a
model-agonostic meta-learner via double-gradient update to
provide an effective prior for task-specific base-learners.
Then, various MAML variants stabilize the system, and
improve the generalization performance, convergence speed
and computational overhead \cite{12,22}.

MAML consists of two levels of optimization, inner and
outer. Inner optimization updates parameters of the model
adapting to a new task while outer or meta-optimization
learns parameters which are needed for all other training
tasks. There are several meta-learning approaches targeting
the out-of-distribution (OOD) problem. However, their definition
for OOD and the main purpose are different from
ours. \cite{17} introduce OOD-MAML, an extension of MAML,
to detect unseen classes by learning artificially generated
fake samples which are considered to be OOD samples.
They define OOD as samples from unseen classes and this definition is different from our goal of learning invariant
correlations. \cite{21} suggest Bayesian-TAML for imbalanced
and OOD tasks. In Bayesian-TAML, the OOD generalization
means relocating the initial parameters for adaptation to
unseen OOD tasks, because parameters may be less useful
for the adaptation of OOD tasks.

To overcome such distribution shifts, many methods are
proposed to address bias caused by distribution shift for
general machine learning problems. There are mainly three
branches of methods for the selection bias caused by distribution
shift, namely domain adaptation \cite{14,37,31}, distributionally
robust optimization (DRO) \cite{8,7,33,30,16,13} and invariant learning \cite{1,19,3,6}. Domain
adaptation methods aim to reduce bias by learning domaininvariant
representations, which is learned by minimizing a
certain discrepancy between distributions of source and target
features extracted by a shared representation learner \cite{14,37,31}. \cite{14} put forward the domain adversarial neural
network (DANN). A domain discriminator is trained to
distinguish source features from target features and a feature
extractor to confuse the discriminator. Since then, a series
of works have appeared and achieved significantly better
performance. \cite{37} proposed an architecture that employed
asymmetric encodings for target and source data. DRO
methods propose to optimize the worst-case risk within an
uncertainty set, which lies around the observed training distribution
and characterizes the potential testing distributions
\cite{16,13}. However, to better capture the testing distribution,
the uncertainty set should be pretty large in many real-world
scenarios, which also results in the over-pessimism problem
of DRO methods \cite{16,13}. Realizing the difficulty of solving
selection bias caused by distribution shift problem without
any prior knowledge or structural assumptions, invariant
learning methods assume the existence of causally invariant relationships between some predictors $\Phi(X)$ and the target $Y$ \cite{23,1,6}. \cite{1} and \cite{19} propose to learning an invariant representation through multiple training environments.

\section{Method}
Previous meta-learning works focus on in-distribution setting, where the train and test tasks are sampled iid from the same underlying task distribution. For training task $\mathcal{T}_{ {tr }}$ validation task $\mathcal{T}_{ {val }}$ and testing task $\mathcal{T}_{ {te }}$, we have $\mathcal{T}_i \sim P$. This in-distribution setting is highly restrictive. For example, training tasks and testing tasks always have different classes, however, they come from the same identical distribution $P$. In this paper, we consider a more realistic setting: crossdomain setting, where testing distribution is different from training distribution.

\subsection{Robust Meta Learning}
In this section, we introduce our proposed Robust Meta
Learning (RML) to learn invariant representation for imagebased
tasks with distribution shifts.

\textbf{Invariant Risk Minimization}. To learn invariant representation,
previous works propose Invariant Risk Minimization
(IRM), which is intended to estimate optimal invariant causal predictor $f$ by learning correlations invariant across all of the training environments. This also means to find a feature representation of data such that the optimal classifier is invariant across all environments. Based on this idea, IRM is formulated as the following constrained bi-level optimization problem:
\begin{equation}
\begin{aligned}
& \min _{\substack{\Phi: \mathcal{X} \rightarrow \mathcal{H} \\
		\omega: \mathcal{H} \rightarrow \mathcal{Y}}} \sum_{e \in \mathcal{E}_{\mathrm{tr}}} R^e(\omega \circ \Phi) \\
& \text { subject to } \omega \in \underset{\bar{\omega}: \mathcal{H} \rightarrow \mathcal{Y}}{\arg \min } ~R^e(\bar{\omega} \circ \Phi), \forall e \in \mathcal{E}_{\text {tr }}
\end{aligned}
\end{equation}
where $\omega$ is a classifier (or final layer for regression problems), $\Phi$ is a data representation and $\omega \circ \Phi$ is a predictor. However, it is difficult to solve such optimization problem directly since multiple constraints must be solved jointly. \cite{1} introduce IRMv1, a practical version of IRM, with Lagrangian form to relax joint constraints such that the classifier $\omega$ is ``approximately locally optimal''. IRMv1 also assumes linear classifier $\omega$ as fixed scalar. The learning objective of IRMv1 is as follows:
\begin{equation}
\min _{\Phi: \mathcal{X} \rightarrow \mathcal{H}} \sum_{e \in \mathcal{E}_{\mathrm{tr}}} R^e(\Phi)+\lambda \cdot\left\|\nabla_{\omega \mid \omega=1.0} R^e(\omega \cdot \Phi)\right\|^2
\end{equation}
where $\lambda \in(0, \infty]$ controls the invariance of the predictor $1 \cdot \Phi(x)$. The gradient norm penalty $\left\|\nabla_{\omega \mid \omega=1.0} R^e(\omega \cdot \Phi)\right\|^2$ indicates the optimality of fixed linear $\omega=1.0$.

Compared to an ideal IRM objective which possibly regards $\Psi_c$ with sampling noise as non-invariant features, IRMv1 with fixed $\lambda$ is more robust for the sampling noise of finite samples \cite{18}. However, its translated form also introduces severe limitations on the capabilities of ideal IRM objective. The linear fixed classifier assumption of IRMv1 can guarantee to find optimal invariant classifier under the condition that the number of spurious correlations is smaller than the number of training environment, $E>\left|\Psi_s\right|$ \cite{28}. With simple violation of this condition, $E \leq\left|\Psi_s\right|$, IRMv1 can catastrophically fail to generalize for OOD. Furthermore, as the number of finite sample decreases, it is more likely to choose a fake invariant predictor which relies on environmental features, $\Psi_s$ \cite{28}.

\textbf{Robust Meta Learning (RML)} is an optimization-based meta-learning approach, where it learns invariant representaion for different image-based task, which is the main difference from Model-agnostic meta-learning (MAML) \cite{9}. For some parametric model $f_\theta$, RML aims to find a single set of parameters $\theta$ which, using a few optimization steps, can be successfully adapted to any novel task, where the invariant representation is learned. For a particular task, we update our model $f_\theta$ by:
\begin{equation}
\theta_i^{\prime}=\mathcal{G}\left(\theta, \mathcal{D}^{t r}\right),
\end{equation}
where $\mathcal{T}_i=\left(\mathcal{D}^{t r}, \mathcal{D}^{ {val }}\right)$ is the task, $\mathcal{G}$ is gradient descent, and $\theta_i^{\prime}$ is the task-specific model.

With such update rule, the task-specific model $\theta_i^{\prime}$ could not learn invariant representaion, the reason is that the model only predict in the same distribution.

\begin{table*}
	\centering
	\caption{Average accuracy (\%) comparison to state-of-the-arts with 95\% confidence intervals on 5-way classification tasks
		under the conventional setting. Best results are displayed in boldface.}
	\begin{tabular}{ccccccc}
		\hline
		\multirow{2}{*}{Method}&\multicolumn{2}{c}{miniImageNet}&\multicolumn{2}{c}{CUB}&\multicolumn{2}{c}{SUN}\\\cline{2-7}
		&5-way 1-shot& 5-way 5-shot&5-way 1-shot& 5-way 5-shot&5-way 1-shot& 5-way 5-shot\\\hline
		{ Meta-Learner LSTM } & 24.99 & 29.79 & 36.23 & 44.39 & 30.99 & 44.86 \\
		{ MAML } & 45.69 & 60.90 & 48.87 & 63.99 & 57.75 & 71.45 \\
		{ Reptile } & 26.59 & 39.87 & 27.21 & 42.35 & 28.30 & 51.62 \\
		{ Matching Network } & 47.63 & 56.28 & 53.06 & 62.19 & 55.02 & 62.57 \\
		{ Prototypical Network } & 46.15 & 65.56 & 48.21 & 57.80 & 55.70 & 67.32 \\
		{ Relation Network } & 47.64 & 63.65 & 52.76 & 64.71 & 58.29 & 72.15 \\
		{ Baseline } & 23.84 & 32.09 & 25.14 & 35.35 & 27.44 & 34.54 \\
		{ Baseline++ } & 30.15 & 41.19 & 32.48 & 42.43 & 35.56 & 44.42 \\
		{ \textbf{RML} } & \textbf{48.55} & \textbf{66.93} & \textbf{54.53} & \textbf{65.71} & \textbf{59.48} & \textbf{73.42}\\\hline
	\end{tabular}
\end{table*}
\begin{table*}
	\centering
	\caption{Average accuracy (\%) comparison to state-of-the-arts with 95\% confidence intervals on 5-way classification tasks
		under the cross-domain setting. Best results are displayed in boldface.}
	\begin{tabular}{ccccccc}
		\hline
		\multirow{2}{*}{Method}&\multicolumn{2}{c}{miniImageNet$\rightarrow$CUB}&\multicolumn{2}{c}{miniImageNet$\rightarrow$SUN}&\multicolumn{2}{c}{CUB$\rightarrow$miniImageNet}\\\cline{2-7}
		&5-way 1-shot& 5-way 5-shot&5-way 1-shot& 5-way 5-shot&5-way 1-shot& 5-way 5-shot\\\hline
		Meta-Learner LSTM & 23.77 & 30.58 & 25.52 & 32.14 & 22.58 & 28.18 \\
		MAML & 40.29 & 53.01 & 46.07 & 59.08 & 33.36 & 41.58 \\
		Reptile & 24.66 & 40.86 & 32.15 & 50.38 & 24.56 & 40.60 \\
		Matching Network & 38.34 & 47.64 & 39.58 & 53.20 & 26.23 & 32.90 \\
		Prototypical Network & 36.60 & 54.36 & 46.31 & 66.21 & 29.22 & 38.73 \\
		Relation Network & 39.33 & 50.64 & 44.55 & 61.45 & 28.64 & 38.01 \\
		Baseline & 24.16 & 32.73 & 25.49 & 37.15 & 22.98 & 28.41 \\
		Baseline++ & 29.40 & 40.48 & 30.44 & 41.71 & 23.41 & 25.82 \\
		\textbf{RML} & \textbf{41.06} & \textbf{56.78} & \textbf{48.93} & \textbf{68.53} & \textbf{33.78} & \textbf{43.79}\\\hline
	\end{tabular}
\end{table*}

To learn invariant optimal meta-initialization, we update the model $f_\theta$ with regularization penalty. In meta training phase, the parameters $\theta$ are updated by back-propagating through the adaptation procedure, we have
\begin{equation}
\theta_i^{\prime}=\theta-\alpha \nabla_\theta \mathcal{L}_{\mathcal{T}_i}^{t r}\left(f_\theta\right)
\end{equation}

To learn invariant representation, we have:
\begin{equation}
\mathcal{L}^{ {val }}=\sum_{\mathcal{T}_i \sim p\left(\mathcal{T}^{t r}\right)} \sum_{\mathcal{T}_j \sim p\left(\mathcal{T}^{ {val }}\right)} \mathcal{L}_{\mathcal{T}_j}^{ {val }}\left(f_{\theta_i^{\prime}}\right)
\end{equation}
For environment $e$, we have:
\begin{equation}
\mathcal{L}_e^{ {val }}=\sum_{\mathcal{T}_i^e \sim p\left(\mathcal{T}^{t r}\right)} \sum_{\mathcal{T}_j^e \sim p\left(\mathcal{T}^{v a l}\right)} \mathcal{L}_{e, \mathcal{T}_j}^{ {val }}\left(f_{\theta_i^{\prime}}\right)
\end{equation}
\begin{equation}
\theta \leftarrow \theta-\eta \nabla_\theta \sum_e \mathcal{L}^{e, v a l}
\end{equation}

With such update, we could have instance-specific adapted models $f_{\theta_i^{\prime}}$, which learn invariant representation for tasks.

\section{Experiment}
\subsection{Setups}

We evaluate and analyze our approach in image-based tasks, where tasks are few-shot image classification tasks. Such classification tasks aim at classifying images into $N$ classes with a small number $(K)$ of labeled images, which is named as $N$-way $K$-shot classification. To create imagebased few-shot tasks similar to \cite{10}, multiple widely-used datasets are used, consisting of miniImageNet \cite{38}, CaltechUCSD-Birds 200-2011 (CUB) \cite{40}, and SUN Attribute Database (SUN) \cite{25}. To further validate the effectiveness of our method, we conduct experiments in cross-domain FSL settings. The class set in each dataset is split into three parts without overlap: base classes, validation classes, and novel classes. For the conventional setting, all three parts are from the same dataset. And for the cross-domain setting, validation classes and novel classes are from the same dataset (indicated by the right side of $\rightarrow$), while base classes come from another different dataset (indicated by the left side of $\rightarrow$). Note that validation classes are only used to select the best-trained model. To fit the images from all the datasets to a model, we resize all the images to $84 \times 84$. For comparison, the following representative meta-learning methods are considered on the few-shot image classification task: MAML \cite{10}, Reptile \cite{24}, Matching Network \cite{38}, Prototypical Network \cite{34}, Relation Network \cite{35}, Baseline and Baseline++ \cite{5}.

\subsection{Results}
\textbf{Evaluation Using the Conventional Setting.} The overall
results under the conventional setting are illustrated in Table
1. We observe that our proposed RML method achieves the
best performance for all cases, which indicates that our RML
is robust to different tasks.

\textbf{Evaluation Using the Cross-Domain Setting.} To validate
the effectiveness of our proposed method, we consider a
more challenge setting: cross-domain setting, where the model is testing in the data which has different distribution
from training data. The results are shown in Table 2. In
this more challenge cross-domain setting, all approaches
suffer from a larger discrepancy between the distributions
of training and testing tasks, which results in a performance
decline in all scenarios. However, RML still outperforms all
baselines in all cases, which demonstrate the effectiveness
of our proposed method.

\section{Conclusion}
In this paper, we consider the unknown distribution shift
of the test data. To tackle this challenge, we propose a
novel robust meta learning method, which is more robust
to the image-based testing tasks which is unknown and
has distribution shifts with traning tasks. Our robust meta
learning method can provide robust optimal models even
when data from each distribution are scarce. Experimental
results demonstrate the effectiveness of our proposed method.

\addtolength{\textheight}{-12cm}   % This command serves to balance the column lengths

%%%%%%%%%%%%%%%%%%%%%%%%%%%%%%%%%%%%%%%%%%%%%%%%%%%%%%%%%%%%%%%%%%%%%%%%%%%%%%%%

%%%%%%%%%%%%%%%%%%%%%%%%%%%%%%%%%%%%%%%%%%%%%%%%%%%%%%%%%%%%%%%%%%%%%%%%%%%%%%%%

%%%%%%%%%%%%%%%%%%%%%%%%%%%%%%%%%%%%%%%%%%%%%%%%%%%%%%%%%%%%%%%%%%%%%%%%%%%%%%%%

\end{document}